\documentclass[conference]{IEEEtran}
\IEEEoverridecommandlockouts

\usepackage{cite}
\usepackage{amsmath,amssymb,amsfonts}
\usepackage{algorithmic}
\usepackage{graphicx}
\usepackage{textcomp}
\usepackage{xcolor}

\usepackage{cite}

\usepackage{orcidlink}

\usepackage{latexsym}
\usepackage{amssymb}
\usepackage{amsmath}
\usepackage{amsthm}
\usepackage{booktabs}
\usepackage{enumitem}
\usepackage{graphicx}
\usepackage{color}

\usepackage{xcolor}
\usepackage{comment}
\usepackage{float}

\usepackage[T1]{fontenc}

\usepackage{multirow}
\usepackage{multicol}
\usepackage{booktabs}
\usepackage{arydshln}
\usepackage{hyperref}

\usepackage{subcaption}

\IEEEoverridecommandlockouts
\IEEEpubid{\makebox[\columnwidth]{
\vbox{\raggedright 
\copyright2024 IEEE\\
Published in IEEE ICTAI 2024,
DOI: \href{https://doi.org/10.1109/ICTAI62512.2024.00104}{10.1109/ICTAI62512.2024.00104}}\hfill
} \hspace{\columnsep}\makebox[\columnwidth]{ }}

\begin{document}




\title{A Chatbot for Asylum-Seeking Migrants in Europe\\
\thanks{Bettina Fazzinga, Elena Palmieri, and Margherita Vestoso contributed equally.
Corresponding author: Andrea
Galassi.
This work was partially supported by the following projects:
European Commission's NextGeneration EU programme, PNRR -- M4C2 -- Inv. 1.3, PE00000013 - ``FAIR - Future Artificial Intelligence Research'' -- Spoke 8 ``Pervasive AI’’;
EU H2020's ICT48 project ``Humane AI Net" Grant \#952026; MUR project PRIN 2022 EPICA "Enhancing Public Interest Communication with Argumentation" (CUP H53D23003660006) funded by Next Generation EU.}
}

\author{\IEEEauthorblockN{
Bettina Fazzinga\IEEEauthorrefmark{1}\orcidlink{0000-0001-8611-2377}, 
Elena Palmieri\IEEEauthorrefmark{2}\orcidlink{0000-0001-5176-8843}, 
Margherita Vestoso\IEEEauthorrefmark{3}\orcidlink{0000-0001-7725-8569}, 
Luca Bolognini\IEEEauthorrefmark{2}\orcidlink{0009-0001-9901-575X},\\ 
Andrea Galassi\IEEEauthorrefmark{2}\orcidlink{0000-0001-9711-7042}, 
Filippo Furfaro\IEEEauthorrefmark{1}\orcidlink{0000-0001-5145-1301}, 
Paolo Torroni\IEEEauthorrefmark{2}\orcidlink{0000-0002-9253-8638}
}
\IEEEauthorblockA{\IEEEauthorrefmark{1}University of Calabria, Rende (CS), Italy\\
Email: \{bettina.fazzinga, filippo.furfaro\}@unical.it}
\IEEEauthorblockA{\IEEEauthorrefmark{2}DISI, University of Bologna, Bologna, Italy\\
Email: \{e.palmieri, luca.bolognini6, a.galassi, p.torroni\}@unibo.it}
\IEEEauthorblockA{\IEEEauthorrefmark{3}University of Naples Federico II, Naples, Italy\\
Email: margherita.vestoso@unina.it}
}

\maketitle

\IEEEpubidadjcol



\begin{abstract}
We present ACME: A Chatbot for asylum-seeking Migrants in Europe. ACME relies on computational argumentation and aims to help migrants identify the highest level of protection they can apply for. This would contribute to a more sustainable migration by reducing the load on territorial commissions, Courts, and humanitarian organizations supporting asylum applicants. We describe the background context, system architecture, underlying technologies, and a case study used to validate the tool with domain experts.
\end{abstract}


\section{Introduction}

Addressing migratory phenomena sustainably and equally is universally understood as a crucial global challenge. 
So much so, that the United Nations' 2030 Agenda for Sustainable Development includes, among its targets, to \textit{facilitate orderly, safe, regular and responsible migration and mobility of people}.\footnote{\url{https://www.migrationdataportal.org/sdgs?node=10}}
Not surprisingly, migration is one of the top current concerns of the European Union too, and there's every indication that the phenomenon is expanding.
In 2023 there were almost one million asylum applications in the Member States of the European Union,\footnote{\url{ https://ec.europa.eu/eurostat/data/database}} a 21.3\% increase compared to the previous year and the highest number since 2016. Because of recent conflicts, the number of irregular arrivals and asylum applications is estimated to grow further.\footnote{International Centre for Migration Policy Development (ICMPD), Migration Outlook 2024, \url{https://www.icmpd.org/about-us/icmpd-migration-outlook}.} 
To handle the phenomenon more effectively and humanly \cite{hailbronner2024immigration}, the EU Commission has set different reforms. Among them is the recent Pact on Migration and Asylum, i.e., a set of new rules managing migration and establishing a common asylum system at the EU level. Such a normative effort was made necessary by the complexity of current asylum rules. The application of such rules requires a non-trivial evaluation of legal and factual data concerning the applicant’s living conditions, both in the host country and in the country of origin, which makes it difficult for the territorial commissions, Courts, and even for migrants, to achieve a preliminary overview on their chances to obtain protection.

In this context, a preliminary assessment and support tool would be beneficial in orienting applicants and helping them understand the possibilities and the requirements before they begin the formal asylum request process. The simplest tool could be a checklist. However, that would help with a well-identified simple procedure, but it could hardly tackle the additional, more complex problem of 
understanding which procedures are relevant. Indeed, there is not only one type of protection but several ones.
Importantly, since applicants may be political refugees and victims of abuse, discrimination, and persecution, the collection and processing of their personal data for immigration purposes must satisfy requirements of privacy. Finally, the legal domain requires transparency and auditability concerning the application of the reference normative framework.



In the last years, Large Language Model (LLM)-based chatbots have become a popular choice for non-trivial question-answering systems of all sorts, and there's ample choice of technologies and tools for building systems with an unprecedented level of fluency and a seeming ability to perform basic inference steps. However, it also became apparent that LLM-based chatbots have significant limitations in terms of hallucinations, biases, jailbreaks and anonymity~\cite{review-chatbots,DBLP:conf/ictai/Azaria23}. These limitations undermine their reliability and usability in domains with strong requirements in terms of transparency, auditability, privacy, and where the factuality of the system's answers must be guaranteed. A popular approach to mitigating part of these problems is resorting to Retrieval-Agumented Generation (RAG) architectures~\cite{pmlr-v119-guu20a}, whereby LLMs are prompted to generate outputs only based on verified, traceable sources. Nevertheless, even state-of-the-art RAG systems are hampered by their apparent inability to perform complex (multi-hop) reasoning and integrate information from multiple documents~\cite{chen2024benchmarking}. This is certainly an issue in the domain at hand, where the complex legislation on international protection cannot be broken down into simple independent fragments where answers can be readily found but, on the contrary, answers require complex reasoning from multiple texts.


An alternative approach to building such a tool could rely on computational argumentation. 
The tool we propose builds on~\cite{iswa22chatbot} and is named ACME: A Chatbot for asylum-seeking Migrants in Europe. 
We shall stress that, like~\cite{iswa22chatbot}, whose aim is to inform about COVID-19 vaccines and not play the role of a medical doctor, the purpose of ACME is to support, inform, and guide asylum applicants, not to assist or replace a judiciary expert in a ``predictive justice'' fashion, a topic imbued with ethical issues~\cite{lazar2023predictivejustice} which is outside our scope.
Compared to~\cite{iswa22chatbot}, ACME has a similar system architecture and preserves its properties, but it offers more modularity, an enhanced natural language interface, and can handle more complex domains.
Relevant properties ACME exhibits include: data governance and privacy thanks to its modular architecture; transparency and explainability thanks to argumentative reasoning; and the ability to integrate and reasoning with explicit, expert-made, formalized knowledge, ensuring auditability.

In the next section, we provide the essential background. We then present the chatbot architecture and technologies, and discuss its validation via a small-scale study with experts.
%
%


\section{Background}

\subsection{Legal background}
The \textit{Refugee Status} is the strongest level of protection that can be granted.
It is given to anyone who fears persecution for reasons of race, religion, nationality, membership to a particular social group or political opinion and is unable or unwilling to avail himself of the protection of his country.
\textit{Subsidiary Protection Status} is a second-level strong guarantee which takes place mostly for the same reasons as the refugee status but only if the latter cannot be recognized. Given the similarity of the requirements, for the purposes of this case study, we consider them to be the same.
Finally, \textit{Special} (or \textit{Humanitarian}) \textit{Protection} is a National type of protection; this is a weaker safeguard that is guaranteed to applicants who are not eligible as refugees or beneficiaries of subsidiary protection but worthy of protection due to special humanitarian reasons.
For each kind of protection, the application is grounded on requirements specified by the law, considered also in terms of living law. In most cases, migrants do not know the regulatory framework of international protection; therefore, they ignore which kind of protection they can apply for. Most of the time, they apply for all types although they can only get one, or none.
ACME is meant to help the applicant understand the level of protection they could realistically apply for. To provide such information, ACME uses criteria defined by legal experts that consider both positive discipline and relevant jurisprudence. This kind of data is usually unavailable online but requires lawyers to use on-licence software and databases to get it and then make their evaluations. The knowledge arising from such a process is thus inaccessible to traditional LLMs.

\subsection{Argumentation-enabled chatbots}

Recently there have been some proposals for argumentation-enabled chatbots supporting persuasive interaction~\cite{hunter-covid} and privacy-preserving medical advice~\cite{iswa22chatbot}. Argumentation is particularly suited to this sort of application due to its amenability to capture rich reasoning patterns and interconnections between alternative options and supporting elements, as well as dialogical interactions~\cite{Modgil2013}. Computational argumentation frameworks are ideal candidates to support reasoning tools that must exhibit transparency and auditability requirements, because they are symbolic and thus interpretable~\cite{ArgumentationInAI-book}.

Chatbots can benefit from computational argumentation in terms of explainability and auditability, improving user trust, and exploiting a reasoning engine to guide the conversation. The application of argumentative chatbots spans several topics such as healthcare \cite{erqbot,sassoonexplainwellness} and COVID-19 \cite{hunter-covid,iswa22chatbot,CLARchatbot,NL4AIchatbot}, ethics \cite{hauptmann2024}, university fees \cite{hunter-persuasive,hadoux2021} and education \cite{guoargumate}.  In \cite{hunter-persuasive}, the authors utilize a crowd-sourced argument graph as a knowledge base to support a chatbot persuading the user, considering the UK's university fees as case study \cite{hunter-persuasive}, and then extend the same approach to COVID-19 vaccines \cite{hunter-covid}.
The chatbots described in these studies address the user's concerns by associating them with an argument in the graph and replying with its counterargument.
Aiming at broadening the user's viewpoint, the authors of \cite{hauptmann2024} have developed a German argumentative chatbot to converse about AI-related ethically sensitive topics. The system determines the user's stance on a certain topic and challenges it.
In their study to implement a healthcare chatbot, the authors of \cite{erqbot} employ an Explanation-Question-Reply (ERQ) argument scheme modelling the agents' interactions and clinically specialized argument schemes (ASPT). The chatbot aims to deliver patients' treatment plans and answer the user's questions.
Our work builds on \cite{iswa22chatbot,CLARchatbot}, where the authors propose a modular architecture based on similarity between embeddings and computational argumentation to address a case study on COVID-19 vaccines. One of the shortcomings of such work is that natural language understanding is addressed only through sentence embedding similarity. The ability of the system to understand is therefore tied to the richness of the sentences in the knowledge base and may hinder the comprehension of concepts that are not explicitly expressed in a similar way. Besides the natural language module, other important differences from previous work are the domain, extent, and complexity of the case study, since the COVID-19 case study was much simpler and self-contained, and the implementation, which \cite{iswa22chatbot,CLARchatbot} don't have, whereas ACME has been implemented into an open-source tool.\footnote{\href{https://github.com/lt-nlp-lab-unibo/ACME-A-Chatbot-for-Migrants-in-Europe}{https://github.com/lt-nlp-lab-unibo/ACME-A-Chatbot-for-Migrants-in-Europe}}

\subsection{Chatbots for immigration}

As far as chatbots in the immigration domain are concerned, the authors of \cite{kotiyal2022immigration} leverage a knowledge graph containing information about the UK's immigration laws to aid the user in preparing a visa request using question templates as a language interface, but does not use argumentation in any way.
In \cite{DBLP:conf/ACMdis/ChenLNL20}, the authors conducted a series of activities with migrants and other stakeholders to co-design a personality-driven chatbot aimed at favor integration, focusing on aspects such as the characteristic of the avatar and how to motivate users to train the chatbot.

\section{Architecture and Technology}

ACME is a modular system, with several interfaces to fit different contexts. For example, for practical reasons, we believe a speech-to-text interface will be part of it.
However, this paper focuses on the core components of ACME responsible for reasoning and interaction, and the interplay between computational argumentation and natural language interface.

Like \cite{iswa22chatbot}, ACME has a neuro-symbolic architecture, where a neural module dedicated to understanding the user is combined with a symbolic module dedicated to reasoning. It includes:

\begin{itemize}
    \item 
    a \textbf{knowledge base} (KB) consisting of a text file representing an argument graph and a JSON file that associates a set of natural language sentences with each node;
    \item a neural \textbf{natural language} module whose task is to interact with the user, understanding what they write;
    \item a symbolic \textbf{argumentation} module to reason from all the available data, compute answers and provide explanations. 
\end{itemize}

The user interacts with the natural language module providing relevant information and answering questions.
The information thus elaborated is mapped to KB concepts.
The argumentation module performs reasoning on the KB and finds the appropriate answer. In case a definite answer is not possible, the system identifies a node that would help determine an answer and asks the user to provide such information.

The architecture design is focused on decoupling between the natural language module and the argumentation module.
The latter has access only to the information mapped into the concepts of the KB, which by definition are relevant for inferring an answer. Any other irrelevant information provided by the user is discarded, guaranteeing their privacy.

\subsection{User Interface}

\begin{figure*}[ht]
    \centering
    \begin{subfigure}[l]{0.65\textwidth}
        \centering
        \includegraphics[height=7.5cm]{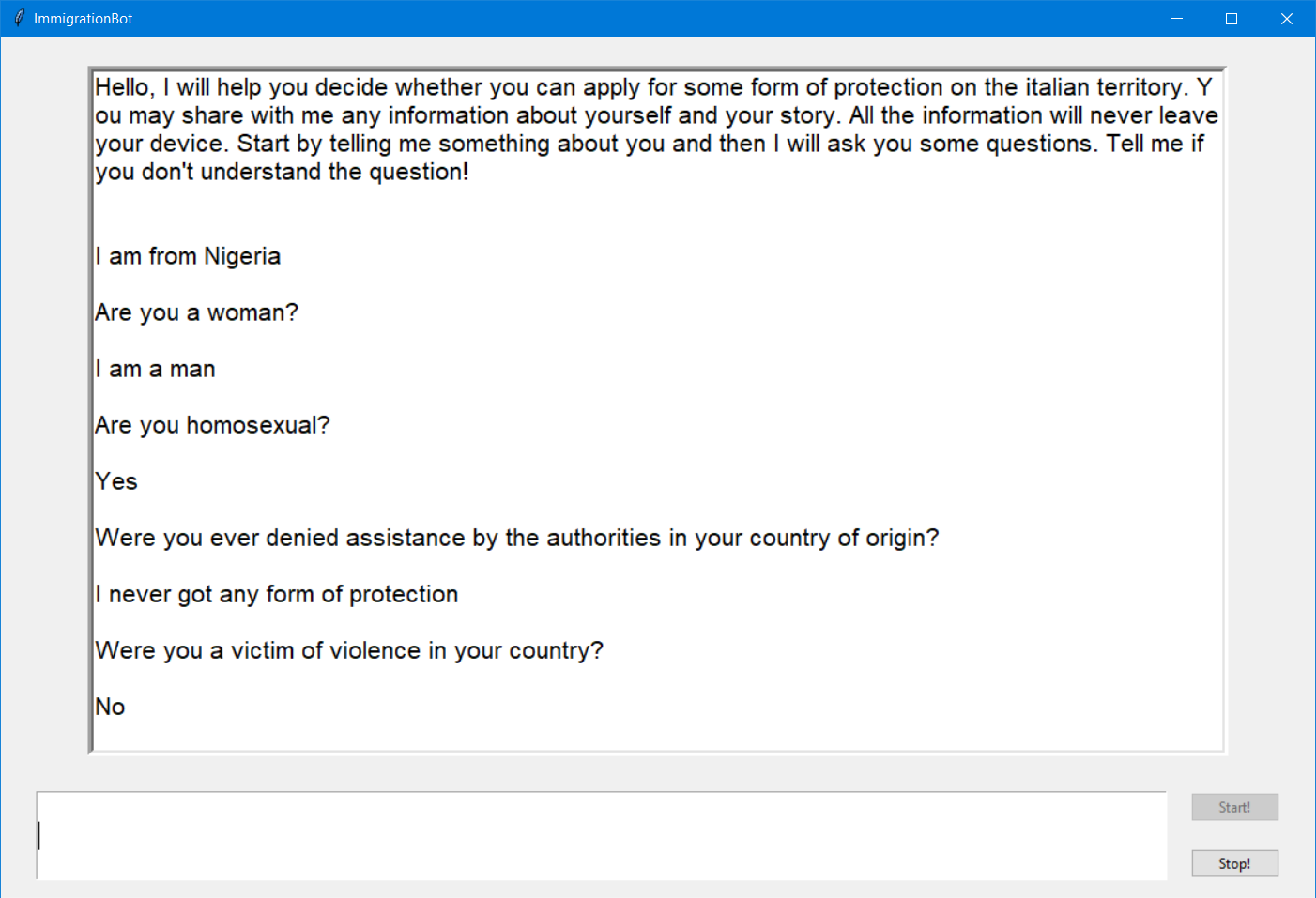}
        \caption{Main chat window}
        \label{fig:interface_main}
    \end{subfigure}
    \hfill
    \begin{subfigure}[r]{0.3\textwidth}
        \centering
        \includegraphics[height=5cm]{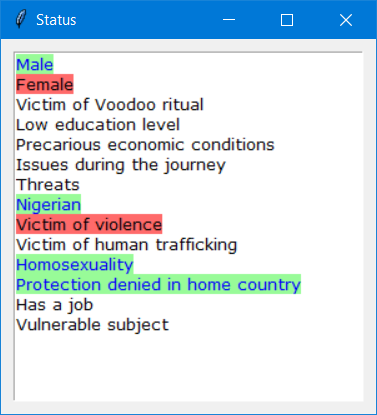}
        \caption{Status window}
        \label{fig:interface_arg}
    \end{subfigure}
    \caption{Overview of the interface.}
    \label{fig:interface_overview}
\end{figure*}

The user interface consists of two  windows: the main chat window and the status window. The main window, represented in Figure \ref{fig:interface_main}, has a standard chatbot layout, with the main dialogue area above and the user input area below.
The status window, in Figure \ref{fig:interface_arg}, contains a list of all the argumentative nodes of the graph.
The nodes are highlighted in real time while the user shares new information. If the node tallies with the user input, the interface will highlight it in green; if it contradicts it, the interface will highlight it in red. 
The UI and the other ACME modules are all implemented in Python.  %

\subsection{Natural Language Module}

The focus of this module is to elicit information from the user and then perform Natural Language Understanding (NLU) to map them into the nodes of the KB.
In particular, the matching of the user answer with KB nodes is performed in several ways.

If the chatbot is inquiring about the information represented in a specific node and the answer of the user is a direct ``yes'' or ``no'', the corresponding node is considered matched.
    
Each node in the KB is associated with a set of natural language strings representing that concept. The user's input and the KB strings are embedded using SentenceTransformers \cite{reimers-gurevych-2019-sentence}. Then, the input is compared against each string and if their similarity is above a fixed threshold, the sentence's node is considered matched.

Otherwise, if the similarity does not reach the threshold, the input is given to a Large Language Model.
Through the use of LLM, the system can understand concepts that are not explicitly presented as they are in the knowledge base, but can still be inferred.
Presently, ACME integrates Llama 2 \cite{touvron2023llama} 70B-chat; however, it is safe to assume that realistically field applications will rely on much smaller language models such as Microsoft's Phi-3.
The LLM is given a prompt that includes the sentence provided by the user, a list of all the nodes of the graph with their textual description, and instructions to return the relevant node.
If all three methods do not find any corresponding node, the chatbot asks the user for clarification.

\subsection{Argumentation Module}
\begin{figure}[ht]
      \includegraphics[width=\columnwidth]{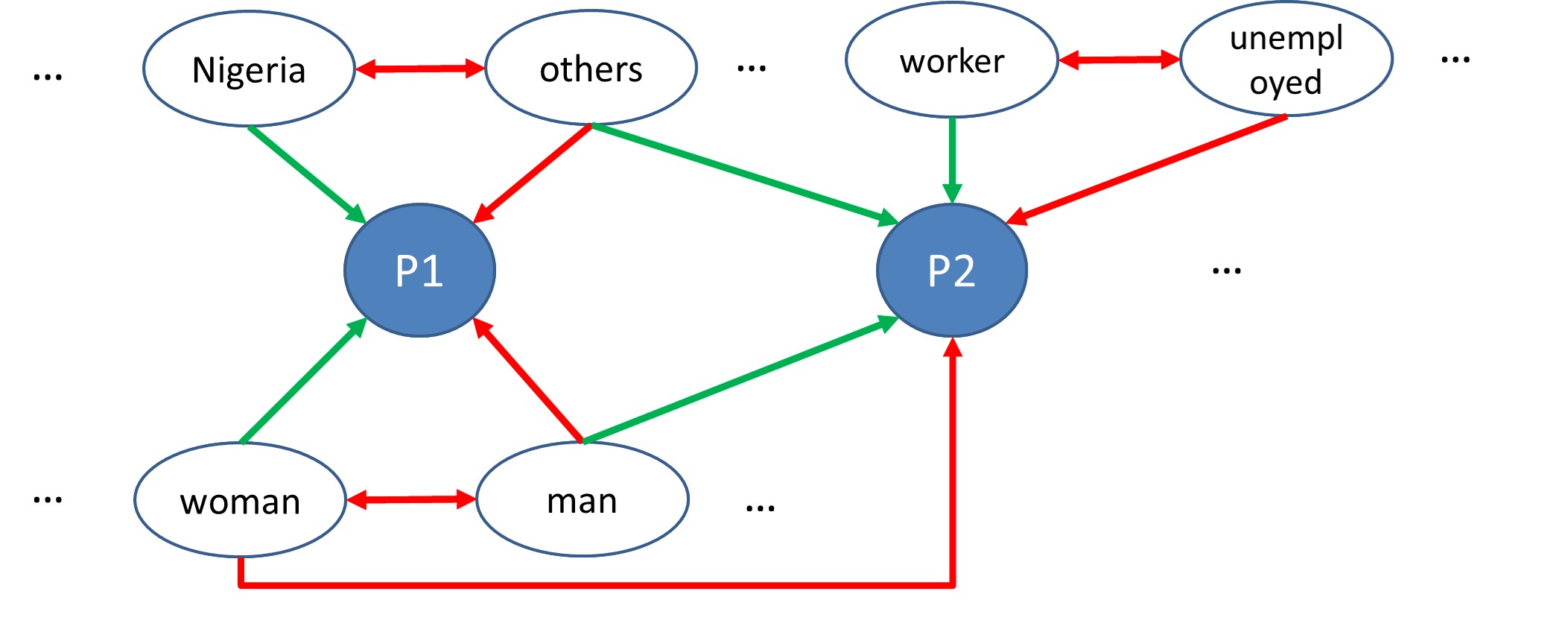}
        \caption{An excerpt of our Knowledge Base}
        \label{fig:KB}
\end{figure}

The Argumentation Module is in charge of selecting the proper questions to
be asked to the user and of computing the replies. This module relies on a Knowledge Base
on the form of an Abstract Argumentation Framework \cite{ArgumentationInAI-book}, where all the facts needed to obtain the different protections are encoded as special arguments named \textit{status arguments}, 
while the different form of protections are encoded as special arguments named \textit{reply arguments}. 
Figure \ref{fig:KB} show an excerpt of our Knowledge Base. Arguments $P_1$ and $P_2$, depicted in blue in the figure, are the \textit{reply arguments}, as $P_1$ and $P_2$ encode the different forms of protections. All the other arguments,  depicted in white, are the \textit{status arguments}.

Each status argument is linked to one or more reply arguments it \textit{endorses}. Status nodes may also \textit{attack} other status or reply arguments, typically because the facts they represent are incompatible with one
another. In Figure \ref{fig:KB}, the endorsement relation is depicted in green, while the attack  relation is depicted in red. For example, argument \emph{woman} mutual attacks argument
\emph{man}, endorses $P_1$ and attacks $P_2$, because $P_1$
is a form of protection intended for women and $P_2$ is
a form of protection destined for men.


Each dialogue session relies on dynamically
acquired knowledge, expressed as a set of status arguments $S$, that encode
user information. Basically, $S$ contains the status nodes of the KB \textit{activated}
so far, that is corresponding to the information the user has communicated
to the system since the start of the dialogue session.
Differently from other proposals, at each turn, our system does not simply
select a reply endorsed by $S$. On the contrary, the aim of the dialogue strategy
is to provide the user with a reply that is both endorsed and \textit{defended} by $S$ from
every attack. In other words, the system works to provide only robust replies, possibly
delaying replies that need further fact-checking. 
In fact, our system distinguishes between \textit{consistent} and \textit{potentially consistent} reply. The former can be given to the user right away, as it can not possibly be proven wrong in the future. The latter, albeit consistent with the current known facts, may still
be defeated by future user input (that means that it is attacked by some argument and it
is not currently defended by what it is in $S$), and therefore it should be delayed until a
successful elicitation process is completed.
Basically, at each turn, the module searches for a consistent reply, and if it does not exist, a potentially consistent reply is selected and all the arguments that can lead to its defense are tried to be added to $S$ by asking the user the proper questions. 
For example, supposing that the chatbot has acquired  the information that the user is a woman, $S$ contains the \emph{woman} argument, but ACME can not give right away the $P_1$ outcome to the user, as $P_1$ is attacked by the \emph{others} argument (and by many others in the actual KB). Thus,
$P_1$ is a \textit{potentially consistent} reply, and the chatbot has to ask more questions to the user in order to promote $P_1$ to a \textit{consistent} reply. To this end, ACME has to get to know, as an example,
whether the user comes from Nigeria: in this case, the attack to $P_1$ from argument \emph{others} is ``neutralized'' by argument \emph{Nigeria}, as the latter attacks \emph{others} and, thus, \textit{defends} $P_1$.

Besides providing replies, the system can also provide \textit{explanations} for the given replies. An explanation of a reply $r$ consists of two parts.
The first one contains the arguments leading to $r$, i.e., those belonging to the
set $S$ that endorses $r$. The second one encodes the \textit{why nots}, to explain why
the system did not give other replies.

\section{Validation}

We implemented an asylum request case study in collaboration with immigration lawyers and other domain experts, based on an analysis of data on asylum applications in Italy over the last 10 years. We focus on the possible use of ACME when applying for international protection in Italy by people whose country of origin is Nigeria,
considering the conditions required by two protection classes: the Refugee/Subsidiary Protection Status, and the Special Protection.

\subsection{Modeling}
We defined 13 argument nodes, each with its opposite, for a total of 26, and a node for each possible reply, namely Refugee Status, Special Protection and no protection. The argumentative nodes are all relevant to understand the person's background and potentially grant them one of the possible protections. They include gender and country of origin, and a collection of nodes representing all the possible reasons that motivate and support some form of protection. These include being a victim of human trafficking, violence, or threats, being employed in the hosting country, being homosexual, having a precarious economic condition, a low level of instruction, being vulnerable. For example, someone who has to provide for their family or has children born in the hosting country is considered vulnerable. 

Each node attacks its opposite node. For example, if a person states that he is a man, he cannot state that he is a woman. If he states that he is from Nigeria, he cannot state that he is from another country. If something like that happens during the conversation, ACME will recognize the contradiction thanks to the attack relation between the opposite nodes and ask the user for clarification. 

Other attack relations are between argument nodes and reply nodes, for example, if someone lacks the criteria of vulnerability, it is unlikely that they will get the Special Protection, so there is an attack relation between the argumentative node ``No Vulnerability'' and the ``Special Protection'' reply node. On the other hand, the vulnerability condition can lead to the Special Protection, so there is a support relation between the two nodes. Another support relation exists between the non-existence of conditions that can lead to the strongest form of protection, i.e., the Refugee Status, and the weaker Special Protection node.

\subsection{Experiment}
To validate our tool on this case study, we run a double-blind experiment involving a human expert on the domain.
First, we generated 10 cases that simulate an asylum-seeking person (profiles). We did so, by assigning a random value to each relevant element considered in our knowledge base (such as country of origin, employment situation, etc.).
For each case, we interacted with the chatbot by impersonating a possible applicant, coherently with the randomly generated profile, describing the case through natural language and eventually obtaining an answer.
We then provided the same facts in the profile to the legal expert, and asked them to determine the highest degree of asylum the applicant could request. For the sake of this experiment, the expert was instructed to consider all the provided elements as true,
independently of how plausible they might be.
Finally, we compared expert answers with ACME answers.

Our tool and the legal expert agreed on all 10 cases, giving the same answer, thus validating our work for this specific case study.

For comparison, we run the same experiment with a state-of-the-art LLM with zero-shot prompting. We instructed GPT-4o to impersonate an expert in Italian immigration law tasked with aiding an asylum-seeking migrant to identify the strongest form of protection the migrant can apply to. We gave the migrant's statements as they were listed in the relevant profile, and instructed the model to choose between the options Special Protection, Refugee Status and no protection, and to justify the answer. We tried with two different prompts. In both cases, GPT-4o returned the correct answer only 3 times out of 10. This corroborates our hypothesis that even the most advanced LLMs do not have access to the relevant jurisprudence, or are unable to use the facts they know meaningfully in this domain. In two cases, the same questions asked using different prompts yielded different answers, showing a fundamental lack of robustness. We also found it concerning that the argumentation used by the LLM to justify the (mostly random) answers did sound like expert advice. 

\section{Discussion}
It has often been observed that legal knowledge is difficult, if not impossible, to access for the individuals who should benefit from it~\cite{TheForceAwakens}. The point has been made in particular for consumer law. However, it also holds for the regulatory framework of international protection: a system whose beneficiaries, the migrants, are usually unable to navigate alone.
In this work we described a system, ACME, intended to facilitate a sustainable approach to migration by guiding and supporting a migrant or someone acting on their behalf towards identifying the highest level of protection they can apply for. 

We are aware of possible ethical issues arising from the adoption of AI tools in the legal domain \cite{lazar2023predictivejustice}. However, ACME is not meant to assist or replace international protection judges. Indeed, a Court's decision to approve or reject an asylum application is grounded on evaluations, like assessing the credibility and coherence of the applicant's narration or the usefulness of the proof provided, which go well beyond the scope of our tool.
For example, during the validation of the case study, the legal expert pointed out that one of the cases presented a suspicious degree of incoherence. An asylum application based on the same elements would have been probably regarded as not credible and eventually rejected by a court. These considerations regarding the truthfulness of the applicant's declarations are outside the scope of this work.  Our purpose is to support migrants before they present asylum applications to inform and guide them with such a complex procedure, and as such, it has to be regarded as an effort to make migration more sustainable.
When used in the field, ACME will prominently exhibit a disclaimer to that effect.

In its present realization, ACME is only a proof-of-concept prototype. However, it is the first one of its kind: a necessary step towards exploring the suitability and potential of the interplay between computational argumentation and natural language processing in such a complex domain. We hope that our work can inspire and motivate further research this domain.
%
%
%
In the future, we plan to extend our proof-of-concept to cover a larger variety of situations and to include further modules, such as a speech-to-text interface and automatic translation. 

\section*{Acknowledgment}
We thank Avv. Mariapia Frisina for the valuable insights and discussions.
%


\bibliographystyle{IEEEtran}
\bibliography{IEEEabrv,mybibfile}


\end{document}